\documentclass{ecai} 

\usepackage{latexsym}
\usepackage{amssymb}
\usepackage{amsmath}
\usepackage{amsthm}
\usepackage{booktabs}
\usepackage{enumitem}
\usepackage{graphicx}
\usepackage{color}
\usepackage[dvipsnames]{xcolor}
\usepackage{multirow}

\newcommand{\BibTeX}{B\kern-.05em{\sc i\kern-.025em b}\kern-.08em\TeX}

\begin{document}


\begin{frontmatter}


\paperid{2676} 


\title{GABInsight: Exploring Gender-Activity Binding Bias in Vision-Language Models}


\author[A]{\fnms{Ali}~\snm{Abdollahi}
\footnote{Equal contribution.}
}
\author[A]{\fnms{Mahdi}~\snm{Ghaznavi}\footnotemark
}
\author[A,D]{\fnms{Mohammad Reza}~\snm{Karimi Nejad}
} 
\author[A]{\fnms{Arash}~\snm{Mari Oriyad}}
\author[A]{\fnms{Reza}~\snm{Abbasi}}
\author[A]{\fnms{Ali}~\snm{Salesi}}
\author[B,C]{\fnms{Melika}~\snm{Behjati}}
\author[A]{\fnms{Mohammad Hossein}~\snm{Rohban}
}
\author[A]{\fnms{Mahdieh}~\snm{Soleymani Baghshah}
}

\address[A]{Sharif University of Technology}
\address[B]{ecole polytechnique fédérale de lausanne (EPFL)}
\address[C]{IDIAP research institute}
\address[D]{Shamsipour Technical and Vocational College}


\begin{abstract}
Vision-language models (VLMs) are intensively used in many downstream tasks, including those requiring assessments of individuals appearing in the images. While VLMs perform well in simple single-person scenarios, in real-world applications, we often face complex situations in which there are persons of different genders doing different activities. We show that in such cases, VLMs are biased towards identifying the individual with the expected gender (according to ingrained gender stereotypes in the model or other forms of sample selection bias) as the performer of the activity. We refer to this bias in associating an activity with the gender of its actual performer in an image or text as the Gender-Activity Binding (GAB) bias and analyze how this bias is internalized in VLMs. To assess this bias, we have introduced the GAB dataset with approximately 5500 AI-generated images that represent a variety of activities, addressing the scarcity of real-world images for some scenarios. To have extensive quality control, the generated images are evaluated for their diversity, quality, and realism. We have tested 12 renowned pre-trained VLMs on this dataset in the context of text-to-image and image-to-text retrieval to measure the effect of this bias on their predictions. Additionally, we have carried out supplementary experiments to quantify the bias in VLMs’ text encoders and to evaluate VLMs’ capability to recognize activities. Our experiments indicate that VLMs experience an average performance decline of about 13.2\% when confronted with gender-activity binding bias.
\end{abstract}

\end{frontmatter}


\section{Introduction}
\label{sec:intro}

Nowadays large multi-modal models have shown tremendous potential in various tasks, from information retrieval systems to image captioning, visual question answering, image generation, and visual reasoning.
As a branch of multi-modal models, Vision-Language Models (VLMs)~\citep{pmlr-v139-radford21a}
that provide a shared cross-modal embedding space between text and image modalities, 
have been intensively investigated by researchers recently and are used in many real-world applications~\citep{zhou2023vision, Salin_2023_ICCV, gan2022vision}. 


Despite their wide range of success in downstream tasks, VLMs are subject to many critical biases, which can consequently affect their performance, especially in sensitive applications. For example, \citet{Agarwal2021EvaluatingCT} provides a comprehensive analysis of biases and their implications in the CLIP Models, revealing how this model is exposed to gender biases and how challenging it is to trust it in real-world applications.
As \citet{Srinivasan2021WorstOB} and \citet{DBLP:journals/corr/abs-2309-14381} state, the bias is introduced in visual-linguistic pre-training due to unfairness in the training data, and additionally, at inference time, the visual and linguistic contexts that is used for few-shot applications can also promote bias. 


One important type of bias, which is highly discussed in the literature, is the gender bias. Gender bias demonstrates the undesirable association of a factor with a gender according to the model. For instance, due to the high occurrence of data in which men are repairing devices, this activity is usually associated with men by deep models, which leads to failure when faced with women repairing devices.
~\citet{hall2023visogender} suggests that different VLMs do not perform equally well on determining gender for an occupation, and do not assign equal retrieval likelihood to images of male and female professionals. Also, \citet{DBLP:journals/corr/abs-2309-14381} and \citet{Wang2021AreGQ} show that CLIP~\citep{pmlr-v139-radford21a} shows biased behavior in retrieval based on gender-neutral queries. Please note that, in line with common practice in the literature, our focus is on two genders, masculine and feminine. This approach streamlines the process of generating the dataset and analyzing the experiments.

In this research, we delve into the issue of gender-activity binding in retrieval tasks. We aim to retrieve the corresponding caption or image from two given captions or images, each depicting an activity performed by a different gender, based on a provided image or caption that shows one of the genders performing the activity. We refer to this bias in associating gender and activity as the \textit{Gender-Activity Binding (GAB)} bias. This bias arises due to ingrained gender stereotypes in the model or other types of sample selection biases~\citep{Bareinboim2014RecoveringFS}. Our research indicates that VLMs do not display a substantial bias in retrieval tasks when both text and image modalities depict only one gender. This is because the identification of the performer’s gender can directly lead to the correct output, eliminating the need to bind the performed activity with the performer’s gender. Though, text encoders have shown a considerable bias towards the expected gender. However, The bias becomes more pronounced when the scene is more complex, i.e., when two individuals of different genders are present in the image. We observe a drop in retrieval accuracy in scenarios where the activity performer is the unexpected gender. 
For example, in the context of the “repairing” activity, the retrieval accuracy of VLMs in identifying the performer decreases when a woman is repairing a device and a man is also present in the scene. This is in comparison with scenarios where the performer is a man, or there is no man present in the scene to associate the activity with him.

The significance of this bias escalates when we consider its underlying implications. Not only does it perpetuate societal biases and fairness concerns related to gender stereotypes, but it can also lead to serious complications if incorporated into judgement and decision-making systems. The potential for such bias to inadvertently influence outcomes underscores the importance of addressing it.

\begin{figure*}[ht!]
\centering
\includegraphics[width=0.8\textwidth]{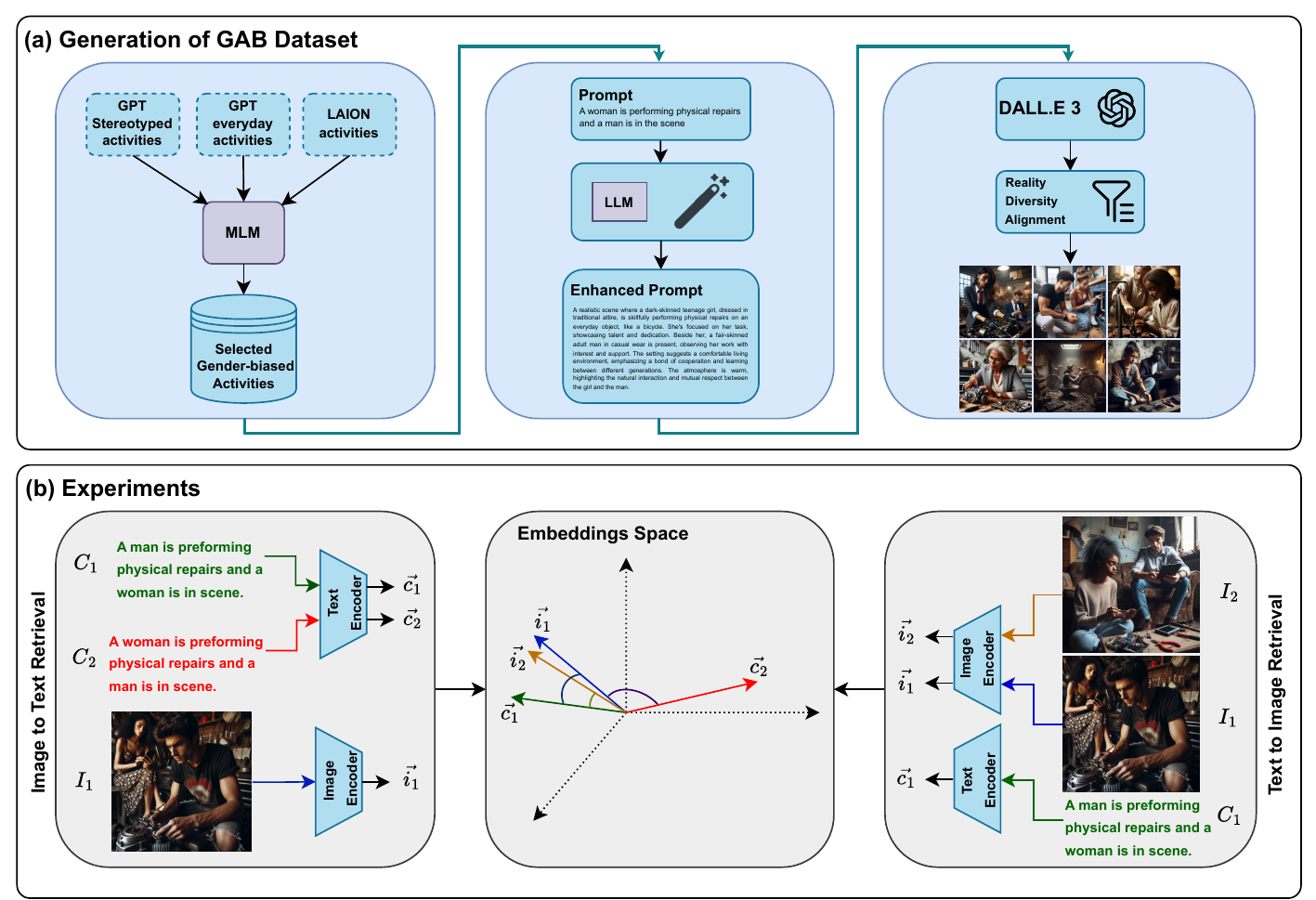}
\caption{Overview of (a) the creation process of the introduced dataset and (b) the empirical tests conducted to assess the gender-activity binding bias in retrieval tasks within vision-language models. (a) Left: we gather three sets of activities that show bias, including stereotypical, everyday activities, and those that exhibit gender bias in the captions of LAION-400M~\citep{schuhmann2021laion400m}. Middle: we employ prompt enhancement techniques to develop a diverse, descriptive, and detailed prompt from a basic initial one, aiding us in generating a wider range of images with superior quality and realism. Right: we utilize DALL-E 3 to construct our dataset based on the enhanced prompts. The generated images are selected to align with the activity and scenario and are evaluated for diversity, quality, and realness to achieve a high score based on standard metrics. (b) Middle: joint embedding space of text and images in vision-language models. Left/Right: an overview of image-to-text/text-to-image retrieval tasks. The caption/image with the highest cosine similarity to the input image/caption is retrieved.}
\vspace{1cm}
\label{fig:main}
\end{figure*}

We have created a dataset known as the \textit{Gender-Activity Binding (GAB)} dataset as shown in Figure~\ref{fig:main}(a). This dataset comprises images depicting various activities performed by men or women in different contexts (with or without the presence of an individual of the opposite gender), each accompanied by a descriptive caption.
Given that our scenarios are uncommon due to biases and stereotypes, suitable real-world images are scarce for evaluation purposes. To address this issue, all images are generated using DALL-E 3\footnote{https://openai.com/dall-e3}. We employ extensive prompt enhancement techniques to ensure the diversity and quality of the generated images. We also evaluated the generated images on diversity, quality, and realism. We performed multiple quantitative assessments consisting of Fréchet Inception Distance (FID) for both quality and diversity assessment, Structural Similarity Index (SSIM) 
and Learned Perceptual Image Patch Similarity (LPIPS) for diversity assessment, and human feedback for a final qualitative assessment to ensure the final results accurately represent the intended activities and their related contexts.
The activities included in GAB dataset are sourced from various places to encapsulate societal gender stereotypes and other activities that are statistically biased towards a specific gender, as observed in the LAION-400M dataset~\citep{schuhmann2021laion400m}.

We benchmark the performance of 12 renowned multi-modal foundation models and assess their performance in terms of text-to-image retrieval, image-to-text retrieval, and recognizing activities on GAB dataset as shown in Figure~\ref{fig:main}(b). 
Specifically, in the image-to-text retrieval task, we observed that while an unexpected gender is doing a stereotypically biased activity, the average performance of VLMs declined by approximately $33.2\%$ due to the presence of the expected gender in the scene. Moreover, most VLMs undergo an average accuracy reduction of approximately $13.2\%$ when encountering gender-activity binding bias. On the other hand, in the text-to-image retrieval task, most VLMs achieve an accuracy of approximately $50\%$, indicating that their performance is nearly random, and they fail to recognize the performer of the activity based on the text.

As far as we are aware, this research is the first to delve into the origins of the association between genders and activities in vision-language models by conducting a thorough analysis both image-to-text and text-to-image retrieval accuracy of VLMs. It encompasses not only straightforward scenarios involving single individuals, where VLMs demonstrate high accuracy for both genders~\citep{howard2024socialcounterfactuals, aneja2018convolutional, srinivasan2022worst}, but also more complex situations with individuals of different genders~\citep{hall2023visogender}. This comprehensive approach allows us to isolate the impact of the complexity of the scenario on performance decline~\citep{Thrush2022WinogroundPV} and more accurately measure the Gender-Activity Binding bias. Additionally, the study carries out supplementary experiments to investigate the bias in VLMs’ text encoders and their proficiency in recognizing activities.\footnote{The dataset and codes are available on Github: https://github.com/sharif-ml-lab/GABInsight}


To summarize, our contributions are as follows:
\begin{itemize}
    \item We have created a novel dataset, known as the gender-activity binding dataset. This dataset includes AI-generated images of various activities that are typically associated with a specific gender, and are excellent in terms of diversity, quality, and realness (Section~\ref{sec:dataset}).

    
    \item We conduct an intensive performance benchmark of well-known vision-language models in retrieval tasks to assess their robustness against the Gender-Activity Binding bias. Our findings reveal that when two individuals of different genders are present, VLMs exhibit a bias towards binding the activity with the gender that is expected to perform it, resulting in a 13.2\% drop in retrieval accuracy. However, in scenarios involving a single individual, these models demonstrate high image-to-text retrieval accuracy. We also demonstrate that VLMs lack the ability to bind gender and activity in text-to-image retrieval tasks (Section~\ref{sec:experiments}). 

    \item We carry out additional experiments to investigate the binding bias in VLMs’ capability to comprehend activities and the bias present in their text encoders (Section~\ref{sec:experiments} and the Appendix). We delve into the insights that can be derived about how VLMs internalize this bias and its impact on the shared embedding space of VLMs (Section~\ref{sec:discussion}).

\end{itemize}


\section{Related Work}
\label{related_works}


\subsection{Vision-Language Models}
VLMs apply contrastive training to bridge the image and text embedding spaces. Using this shared embedding space, VLMs have exhibited a remarkable performance on a notable number of multi-modal downstream tasks such as image captioning, text-to-image and image-to-text retrieval, visual question answering, and text-to-image generation. As \citet{DBLP:journals/corr/abs-2309-14381} and \citet{Li2023MultimodalFM} mentioned, VLMs can be categorized into three main groups based on their architectures: I) Fusion VL-encoders such as FLAVA~\citep{Singh_2022_CVPR} and LXMERT~\citep{tan-bansal-2019-lxmert}, II) Dual-stream encoders like CLIP~\citep{pmlr-v139-radford21a}, GLIP~\citep{Li2021GroundedLP}, and FILIP~\citep{Yao2021FILIPFI}, and III) VL Encoder-decoder including BLIP~\citep{Li2022BLIPBL}.

\subsection{Gender-activity Binding Bias}
Previous works discussed that VLMs have many categories of biases~\citep{DBLP:journals/corr/abs-2309-14381}, especially gender-related ones~\citep{hall2023visogender}. 
For example, \citet{Srinivasan2021WorstOB} studied bias in VLMs as a task of measuring associations between entities and gender in visual-linguistic models using template-based masked language modeling. Moreover, \citet{DBLP:conf/mm/ZhangWS22} measured gender bias in VLMs by comparing the MASK token prediction probabilities of factual and counterfactual samples and concluded that VLMs inherit human stereotypes from the training data.

On the other hand, there exist a number of studies including a restricted assessment of how a VLM understands a particular activity \citep{Thrush2022WinogroundPV, yuksekgonul2023when, Zhao2022VLCheckListEP}. For example, \citet{Thrush2022WinogroundPV} has introduced a small dataset called Winoground containing objects, attributes, and activities to evaluate the ability of VLMs in compositional reasoning. Moreover, \citet{Zhao2022VLCheckListEP} designed VL-CheckList framework to study the capability of VLMs in understanding objects, attributes, and relations containing activities. However, to the best of our knowledge, there is no systematic study on the ability of VLMs in associating an activity with a gender.

\section{Gender-Activity Binding Bias}
\label{sec:bias}


Despite the recent increase in large-scale data collection, it’s important to note that these datasets are not necessarily devoid of biases. These biases can manifest in large datasets derived from real-world distributions, primarily due to sample selection bias.
The issue causes VLMs to have varying levels of accuracy in recognizing the performer of an activity depending on the gender of the individual involved~\citep{DBLP:conf/mm/ZhangWS22}. 
We refer to this bias in associating an activity with the gender of its actual performer in an image or text as \textit{Gender-Activity Binding Bias}. 

Consider an image-to-text retrieval task, where we have two pieces of text $C_1$ and $C_2$, each describing an activity performed by a different gender. We also have an image $I_1$ depicting a person (either a man or a woman) performing an action. The task is to match the image with the correct piece of text (Figure~\ref{fig:main}-(a) left). Also, consider the text-to-image retrieval task, where we have two images $I_1$ and $I_2$, each showing an activity being performed by a specific gender, and we want to retrieve the correct image based on a given piece of text $C_1$ (Figure~\ref{fig:main}-(a) right). The Gender-Activity Binding bias manifests as a decline in the retrieval accuracy of VLMs when they incorrectly identify the gender of an individual performing activities typically associated with a different gender.

Let's denote the probability of a VLM correctly identifying the performer of an activity in a data $i$ (image or text) as $p$, the activity as $a$, and the gender as $g$. Ideally, the probability $p(g|a, i)$ should be equal for all $g$. 
However, because of the co-occurrence of gender $g$ of the performer and the activity in the dataset, which is referred to as sample selection bias~\citep{Bareinboim2014RecoveringFS}, the gender of the performer becomes correlated with the activity. Thus, we often find that $p(g_1|a, i)> p(g_2|a, i)$, where $g_1$ and $g_2$ represent different genders and $g_1$ shows the expected gender for the performer of $a$. It is also 
In simple terms, VLMs tend to incorrectly select the text or image because they are biased towards associating the activity with a specific gender. This is not always accurate, leading to errors in scenarios where the activity is performed by the less expected gender.

The accuracy of models in image and text retrieval tasks is assessed in two scenarios: one where only a person of an unexpected gender (in terms of the bias) is depicted or described in the text or image, and another where both genders are present in the data, but the activity is performed by only one of them. Consequently, the impact of the bias can be observed as a decrease in accuracy in two situations: when the expected gender is absent or present, and when the activity is performed by the unexpected or expected gender. These cases together provide better insights into how these models handle gender-activity associations under different conditions. The first scenario allows us to understand how well the models can recognize and correctly associate an activity with an unexpected gender when the other gender is not present. This helps us gauge the models’ ability to break away from societal stereotypes and biases. The second scenario, on the other hand, tests the models’ ability to correctly identify the performer of an activity when both genders are present, which is a more complex and realistic situation. This indicates these biases affect the models’ performance in real-world scenarios. Together, these cases provide a comprehensive understanding of the models’ strengths and weaknesses in handling gender-activity binding, paving the way for targeted improvements to mitigate the gender-activity binding bias.

\section{Dataset}
\label{sec:dataset}

This section details the creation process of the Gender-Activity Binding (GAB) Dataset, which serves to evaluate various VLMs against the gender-activity binding bias we have identified and VLMs ability to bind activities and performers. This dataset comprises AI-generated images that illustrate a variety of activities with potential subjects of both genders.


The GAB dataset divides images into four distinct groups: In two groups, both genders are present in the scene. In one of these groups, the expected gender is performing the activity while the other gender is just present in the image, and in the other group, the unexpected gender is performing the activity while the other gender is also in the scene.
In the next two groups, only one gender is present in the scene. In one of them, the expected gender is performing the action, and in the other one, the scenario is reversed, with the unexpected gender acting. 

For each activity, with respect to these grouping, we can consider four groups based on the performer of the activity (expected (E) or unexpected (U)) and the number of persons (with different genders) in the scene (1 or 2): E1, E2, U1, U2.


For each image in the dataset, we have templates that are replaced with activity or gender names, such as "man" and "woman," in accordance with the designed experiments. Details of the experiments and these replacements can be found in Section~\ref{sec:experiments}. The template for images containing two genders is: "a \textless{}man/woman\textgreater{} is \textless{}doing activity\textgreater{} and a \textless{}woman/man\textgreater{} is in the scene" and for images containing one gender, the template is: "a \textless{}man/woman\textgreater{} is \textless{}doing activity\textgreater{}". This specific template can also be used for images with two genders.


Each of the mentioned groups contains at least 26 images for each activity, resulting in a total of 5500 images.



This dataset structure facilitates the evaluation of gender-activity binding bias and the ability of VLMs to bind and recognize activities by definition of different retrieval tasks, as you can see in Section~\ref{sec:experiments}.

Pipeline of the dataset creation is shown in Figure~\ref{fig:main}(a).

\subsection{Selection of Activities}
\label{activityTypes}

We proposed three methods to identify the biased activities, all utilizing Masked Language Modeling (MLM) as part of their approach to detecting biased activities. We utilized the 'RoBERTa-large'~\citep{liu2019roberta} model for its superior performance in masked token prediction, which was attributed to its objective function.
The sentences are formatted as "a \textless{}mask\textgreater{} is doing \textless{}activity\textgreater{}".\footnote{\textless{}mask\textgreater{} is the special mask token of the RoBERTa model} We then calculate the probabilities associated with "man" and "woman"
(as well as related masculine and feminine identifiers like "boy" and "girl"), from which we calculate the log ratios, as specified in Section~\ref{stereoSelection}.

\subsubsection{Stereotyped Activities from GPT Combined with MLM}
\label{stereoSelection}
Our first method involved using GPT-4 to request activities stereotypically associated with men or women in society. We provided over 100 activities from GPT-4 and formed sentences by replacing the \textless{}activity\textgreater{} token in the template of 'a \textless{}mask\textgreater{} is doing \textless{}activity\textgreater{}'. We then inputted these sentences into the RoBERTa model and calculated the following equation for the sentences.
\begin{equation}
    bias(sentence) = log\left(\frac{P(mask = man|sentence)}{P(mask = woman|sentence)}\right)
\end{equation}
We applied this methodology separately for activities stereotypically associated with each gender. For men, this resulted in a normal distribution\footnote{all the tests including log ratios formed normal distributions that we tested using Shapiro–Wilk test for normality} with a positive mean ($\mu > 0$), from which we chose the sentences at the rightmost end. Similarly, for women, it also resulted in a normal distribution but with a negative mean ($\mu < 0$), from which we then chose the sentences at the leftmost end.

\subsubsection{Everyday Activities from GPT Combined with MLM}

In our second approach, we used GPT-4 to compile a list exceeding 1,000 sentences related to hobbies, jobs, and everyday activities. We then masked the subjects of the sentences just like in the first method and asked the RoBERTa model to fill the mask. We stored the log ratios of predicted probabilities. The log ratios formed a normal distribution, from which we selected sentences at each end.

\subsubsection{Activities from LAION-400M Dataset Combined with MLM}

In the third method, we used the LAION-400M~\citep{schuhmann2021laion400m} dataset, which consists of 400 million image-caption pairs. We filtered out the NSFW pairs and sampled 20 million pairs randomly. We then extracted the verbs, subjects, and objects of all caption sentences using spaCy~\citep{honnibal_montani_2023_spacy}, a natural language processing toolkit. We counted the number of masculine and feminine subjects for each verb-object pair and calculated their log ratios. This also formed a normal distribution.

We then used the Z-test to determine whether each verb-object pair had a significantly different ratio compared to the mean of all verb-object pairs. This test is appropriate due to the large dataset size of LAION-400M. By checking if differences are statistically significant, we identify biased activities. This method enhances the reliability of our activities by ensuring they are not a result of random variance or outliers. Then, we created sentences with the biased verb-object pairs as mentioned in Section~\ref{stereoSelection}, masked the subject, and submitted them to the RoBERTa model to fill the mask, conducting the same experiment as in the previous methods to select the biased activities.

\subsection{Generation}

The generation process consisted of two phases. The first involved generating a good prompt that helped the image generation system meet the criteria given below, derived from the base sentence. The second involved generating the image from the prompt. The generated images should accurately represent the original activity, look natural and high quality, and be diverse.
\subsubsection{Prompt Enhancement}
We used a Large Language Model (LLM) to generate diverse prompts while also keeping the original activity intact (Details about the utilized LLM are provided in the Appendix.). We then created a list of predefined settings like the environment of the house, the skin color of the person, etc. Although we could have sampled $N$ prompts from each setting to generate diverse images, we chose to create a list of different combinations of these settings and encode them using MiniLM~\citep{wang2020minilm} sentence transformer~\citep{reimers-2019-sentence-bert}. We then clustered them into K\footnote{K = 6 which resulted in sufficient diversity in our experiments} clusters and sampled $\frac{N}{K}$ samples from each cluster to enforce more diversity semantically. We then generated a diverse prompt by instructing Llama2 to generate diverse prompts that maintain the original activity while describing the environment and the settings using the selected samples. Using these prompts, the generated images were significantly improved in quality and detail.

\subsubsection{Image Generation}
We initially experimented with several diffusion models, as mentioned in the Appendix, to generate images. The generated images were realistic and diverse. However, they lacked representation of the intended activity, especially in the context of compositional generation. We then explored DALL-E3~\citep{openai2023dalle} using OpenAI API. This approach resulted in significant improvement to the quality of compositional generation while also keeping the reality and the clarity of the image. We chose DALL-E3 as our final image generation solution.

\subsection{Image Filtering Methodology}
Our image filtering methodology employs a combination of automated metrics and human evaluation to ensure the generated images are of high quality, relevant, and diverse. The process comprises three steps: quality assessment, diversity metrics, and human feedback, detailed as follows:

\subsubsection{Quality Assessment}
The quality of generated images was assessed using the Fréchet Inception Distance (FID)~\citep{heusel2018gans} alongside criteria emphasizing the representation of human figures, as identified by the YOLOv8~\citep{Jocher_Ultralytics_YOLO_2023} and MTCNN~\citep{DBLP:journals/corr/ZhangZL016} models. The FID score aims to measure the similarity between the generated images and a set of reference images from the COCO~\citep{heusel2018gans} dataset, calculated as:

\begin{equation}
FID(x, y) = ||\mu_x - \mu_y||^2 + Tr(\Sigma_x + \Sigma_y - 2(\Sigma_x\Sigma_y)^{1/2}),
\end{equation}
where $\mu_x$, $\mu_y$ are the feature-wise mean of the real and generated images, and $\Sigma_x$, $\Sigma_y$ are the covariance matrices of the real and generated images, respectively. To achieve high-quality images, we initially filtered out images with high Fréchet Inception Distance (FID) scores. After this filtering process, we successfully reduced the mean FID score to 11.9. Furthermore, for an image to pass our quality assessment, individuals must constitute at least 15\% of the detected objects and must have a detectable face, as confirmed by the MTCNN model. 
It’s worth mentioning that the FID scores for both expected (E1 and E2) and unexpected (U1 and U2) categories of images are 12.1 and 11.7 respectively, signifying an acceptable level of quality across all scenarios.

\subsubsection{Diversity Metrics}
To ensure image diversity, we employed the Structural Similarity Index (SSIM)~\cite{1284395} and the Learned Perceptual Image Patch Similarity (LPIPS)~\cite{zhang2018unreasonable} metrics. The SSIM index measures the similarity between two images; in our context, lower scores indicate greater diversity. It is defined as:

\begin{equation}
SSIM(x, y) = \frac{(2\mu_x\mu_y + c_1)(2\sigma_{xy} + c_2)}{(\mu_x^2 + \mu_y^2 + c_1)(\sigma_x^2 + \sigma_y^2 + c_2)},
\end{equation}
where $\mu_x$, $\mu_y$ are the average intensities, $\sigma_x^2$, $\sigma_y^2$ are the variances, and $\sigma_{xy}$ is the covariance. $c_1$ and $c_2$ are constants added to stabilize the division and avoid numerical instability with small denominators. Our target mean SSIM threshold across all image pairs is 0.046. It’s also worth mentioning that the SSIM for the expected (E1 and E2) and unexpected (U1 and U2) categories of images are 0.046 and 0.045, respectively.
This indicates a satisfactory level of diversity in both image categories.

LPIPS measures the perceptual similarity between two images, using deep network features to closely mimic human visual perception. To indicate greater diversity, we aim for higher LPIPS values, setting the mean target for all pairs at 0.66. Also, the LPIPS scores for the expected (E1 and E2) and unexpected (U1 and U2) categories of images are 0.66 and 0.65, respectively.

\subsubsection{Human Feedback}
After quantitative evaluations, we performed a qualitative analysis with human feedback. This involved scrutinizing images for alignment with the desired activities and contexts. Images that did not align with the activities or did not depict the intended subjects were removed. Additionally, images with unrealistic or nonsensical elements, and those appearing abnormal to an external observer were excluded.

Following this step, a list of images that have passed both quantitative and qualitative evaluations is obtained. As a result, the images are diverse, high-quality, and realistic while also meeting our criteria for the generation phase.

\section{Experiments}
\label{sec:experiments}

 We designed several experiments on our proposed GAB dataset to evaluate Vision-Language Models in text-to-image and image-to-text retrieval tasks as shown in Figure~\ref{fig:main}(b). Based on these experiments, we assessed the defined gender-activity binding bias and the ability of VLMs to recognize gender-biased activities. Finally, we evaluated the bias in the text encoders and image encoders of different VLMs separately.

\subsection{Setup}
\label{sec:setup}

\subsubsection{Task Definitions}
The experiments we have designed to assess the performance of VLMs on the aforementioned aspects are grounded in their zero-shot performance on both text-to-image and image-to-text retrieval tasks.


More formally, consider the space of images as $\mathcal{I}$ and the space of captions as $\mathcal{C}$.
Also consider a VLM with an image encoder $E_\mathcal{C}:\mathcal{C}\rightarrow{}\mathbb{R}^D$ and a text encoder $E_\mathcal{I}:\mathcal{I}\rightarrow{}\mathbb{R}^D$, that map images and texts respectively to a shared embedding space, which is shown in the middle box in Figure~\ref{fig:main} (b). We can define matching score function $s:\mathcal{I}\times\mathcal{C}\rightarrow{}\mathbb{R}^{+}$, which measures the similarity between a caption and an image, as the cosine similarity between the embedding of an image $I$ and the embedding of a caption $C$:

\begin{equation}\label{eq:matching_score}
    s(I, C)=\frac{E_\mathcal{I}(I).E_\mathcal{C}(C)}{\lVert E_\mathcal{I}(I) \rVert \lVert E_\mathcal{C}(C) \rVert}.
\end{equation}

Text-to-image retrieval involves finding the image that best matches a given text from a collection of images. Conversely, image-to-text retrieval is about finding the text that best describes a given image. We compute the similarity using the previously defined similarity measure in the upcoming experiments. For the text-to-image retrieval task, we have a caption $C$ and make comparisons between two images: $I$, which shows an image corresponding to the caption $C$, and $I^R$ which corresponds to $C^R$ (obtained by reversing the gender of subjects in $C$). Similarly, for the image-to-text retrieval task, we have an image $I$ and make comparisons between two captions $C$ and $C^R$. 



\subsubsection{Selected Vision-Language Models}
\label{sec:model_selection}

In our experiments, we benchmark 12 VLMs and present the results of various tests.
For comparing the effects of patch size and backbone size, we report the results for 4 CLIP models released by OpenAI~\citep{pmlr-v139-radford21a}, each with different backbones and patch sizes.

Another model selected for evaluation is NegCLIP~\citep{yuksekgonul2023when}, a version of the CLIP-ViT-B-32 model by OpenAI~\citep{pmlr-v139-radford21a} which is fine-tuned on a modified subset of the Visual Genome Dataset~\citep{krishna2017visual} annotations. NegCLIP is fine-tuned to purportedly enhance the base model's performance on tasks requiring relational and activity understanding.

We also report results for several other notable VLMs that have been published in recent years. These include Eva01~\citep{fang2023eva} and Eva02~\citep{EVA-CLIP}, FLAVA~\citep{Singh_2022_CVPR}, ALIGN~\citep{pmlr-v139-jia21b}, COCA~\citep{yu2022coca} and AltCLIP~\citep{chen-etal-2023-altclip}.

\subsection{Results}
This section presents the outcomes of experiments designed to evaluate model performance across various tasks, including Image-To-Text and Text-to-Image retrieval, and activity recognition(Detailed results and descriptions of activity recognition experiments are provided in the Appendix.). Additionally, this section specifically addresses and reports on the bias observed in the text encoder.

\subsubsection{Image-to-Text Retrieval}
\label{sec:ImgToText}

We assess the performance of the models discussed in Section~\ref{sec:model_selection} on the GAB dataset and calculate their accuracy in retrieving the correct caption as outlined in Section~\ref{sec:setup}. We employ a caption template of the form: \textit{“a \textless{}man/woman\textgreater{} is \textless{}doing activity\textgreater{} and a \textless{}woman/man\textgreater{} is in the scene”}. The results for activities that are stereotypically biased are presented in Figure~\ref{fig:phaz3Bias}, while the results for the other two groups of activities are provided in the Appendix. 


\begin{figure}[ht!]
\centering
\includegraphics[width=0.44\textwidth]{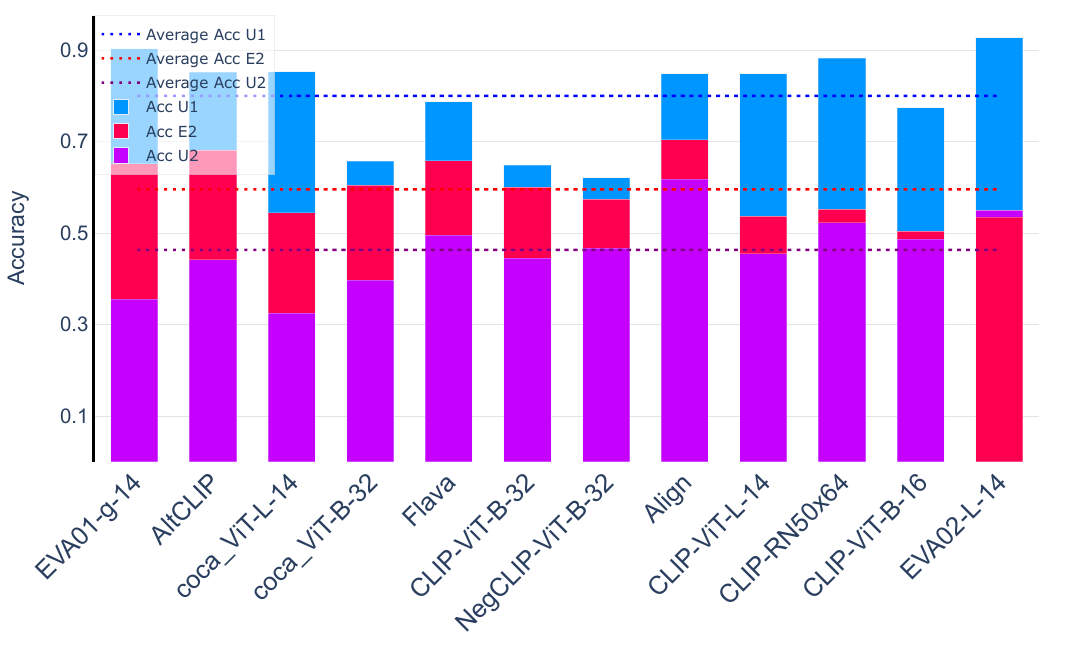}
\caption{Average retrieval accuracy of VLMs on the image-to-text retrieval task across various scenarios. The chart highlights the performance drop between these scenarios for each model. The \textcolor{violet}{purple} bar represents the accuracy in the scenario where the unexpected gender is performing the activity in the reference image, while the expected gender is also present in the scene. The \textcolor{red}{red} bar corresponds to the accuracy in the reverse scenario, where the expected gender is performing the activity. The \textcolor{blue}{blue} bar denotes the scenario where the unexpected gender is performing the activity and is the only one present in the scene.}
\vspace{1cm}
\label{fig:phaz3Bias}
\end{figure}

\begin{figure}[ht!]
\centering
\includegraphics[width=0.44\textwidth]{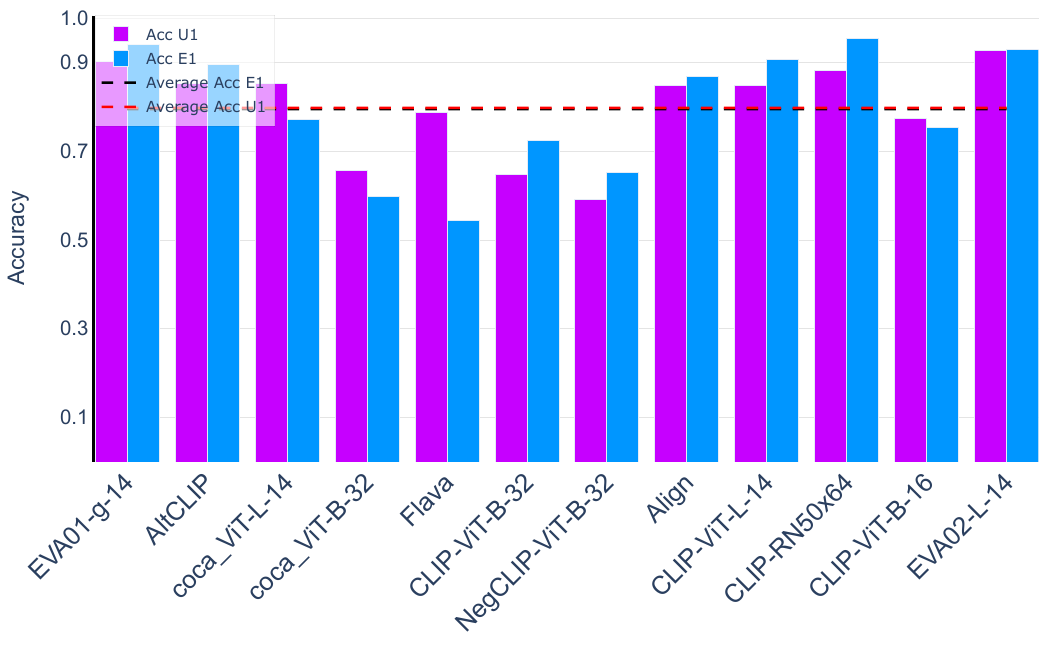}
\caption{Average retrieval accuracy of VLMs on the image-to-text retrieval task on E1 and U1 category of images on stereotypical activities. The \textcolor{violet}{purple} bar represents the accuracy in the scenario where the unexpected gender is performing the activity in the reference image and is alone in the scene. The \textcolor{blue}{blue} bar denotes the scenario where the expected gender is performing the activity and is the only one present in the scene. In this task the template is \textit{“a \textless{}man/woman\textgreater{} is \textless{}doing activity\textgreater{}”}. }
\vspace{1cm}
\label{fig:phaz3Singles}
\end{figure}


In this experiment, we assessed image-to-text retrieval across three distinct scenarios. The first scenario involves images where an individual of an unexpected gender is performing a typically biased activity with no other individuals present in the scene. The second scenario comprises images where both genders are present, and the activity is performed by the gender typically associated with it. The third scenario mirrors the second, but in this case, the activity is performed by the gender not typically associated with it.

\noindent\textbf{Performance Drop Due to Presence of Expected Gender}:
As depicted in Figure~\ref{fig:phaz3Bias}, there is a noticeable drop in the accuracy of the models when the scene includes two genders compared to scenarios where only the unexpected gender is present and performing the activity associated with bias. The chart reveals that most VLMs experience an average performance decline of approximately 33.2\%. However, a few models demonstrate only a slight decrease in performance.

\noindent\textbf{Performance Drop Due to Gender-Activity Binding Bias}: 
In the scenarios where there are two individuals present at the scene, the majority of models display a substantial decline in retrieval accuracy when the action is performed by an individual of the unexpected gender. As indicated in Figure~\ref{fig:phaz3Bias}, the VLMs undergo an average accuracy reduction of approximately 13.2\% when encountering gender activity binding bias. 

Note that, as illustrated in Figure~\ref{fig:phaz3Singles}, the models demonstrate satisfactory performance in the Image-to-Text retrieval task on both image scenarios of U1 and E1, indicating that the gender bias adversely affects the performance of the models only when there is more than one individual in the image. Additionally, the accuracy of the models remains consistent, regardless of whether the expected or unexpected gender is performing the activity. The models' average performance in both scenarios is approximately 80\%.

\subsubsection{Text Encoder Bias}
Given an activity $a$, consider $e,u\in\{\text{man}, \text{woman}\}$ where $e$ is the expected gender of the individual performing $a$ and $u$ is the unexpected one. To separately assess the bias of the text encoder, we determine the frequency at which the embedding of a gender-neutral sentence "a person is \textless{}doing activity\textgreater{}" is closer to the embedding of "a $e$ is \textless{}doing activity\textgreater{}" than the embedding of "a $u$ is \textless{}doing activity\textgreater{}". The outcomes are presented in Table~\ref{Text Bias}. It can be observed that in all categories of the gathered activities, the majority of activities exhibit a bias towards the expected gender.

\begin{table}[ht]
\centering
\caption{Bias of the text encoder for each model. Taking into account the template “a \textless{}mask\textgreater{} is \textless{}doing activity\textgreater{}”, the table shows the proportion of activities where substituting \textless{}mask\textgreater{} with the gender-neutral term “person” results in a higher matching score (Eq.~\ref{eq:matching_score}) in the embedding space to the instance in which \textless{}mask\textgreater{} is replaced by the expected gender reference (woman/man) compared to the unexpected one. The columns denote the collected activities across different categories as outlined in Section~\ref{activityTypes}.}
\vspace{1cm}
\resizebox{0.48\textwidth}{!}{%
\begin{tabular}{lccc}
\toprule
Model & LAION-400M Biased & GPT Gathered & Biased Stereotype \\
& Activities & Biased Activities & Activities\\
\midrule
AltCLIP & 0.80 & 0.73 & 0.91 \\
EVA01-g-14 & 0.80 & 0.80 & 0.92 \\
EVA02-L-14 & 0.80 & 0.80 & 0.92 \\
CLIP-RN50x64 & 0.50 & 0.60 & 0.59 \\
CLIP-ViT-B-16 & 0.70 & 0.60 & 0.83 \\
NegCLIP-ViT-B-32 & 0.70 & 0.60 & 0.75 \\
CLIP-ViT-B-32 & 0.60 & 0.66 & 0.59 \\
CLIP-ViT-L-14 & 0.60 & 0.73 & 0.83 \\
COCA-ViT-B-32 & 0.90 & 0.93 & 0.92 \\
COCA-ViT-L-14 & 1.00 & 0.93 & 0.92 \\
Flava & 0.70 & 0.53 & 0.75 \\
Align & 0.90 & 1.00 & 1.00 \\
\bottomrule
\end{tabular}%
}
\label{Text Bias}
\end{table}

\subsubsection{Text-to-Image Retrieval}\label{sec:ttir}

\begin{figure}[ht!]
\centering
\includegraphics[width=0.44\textwidth]{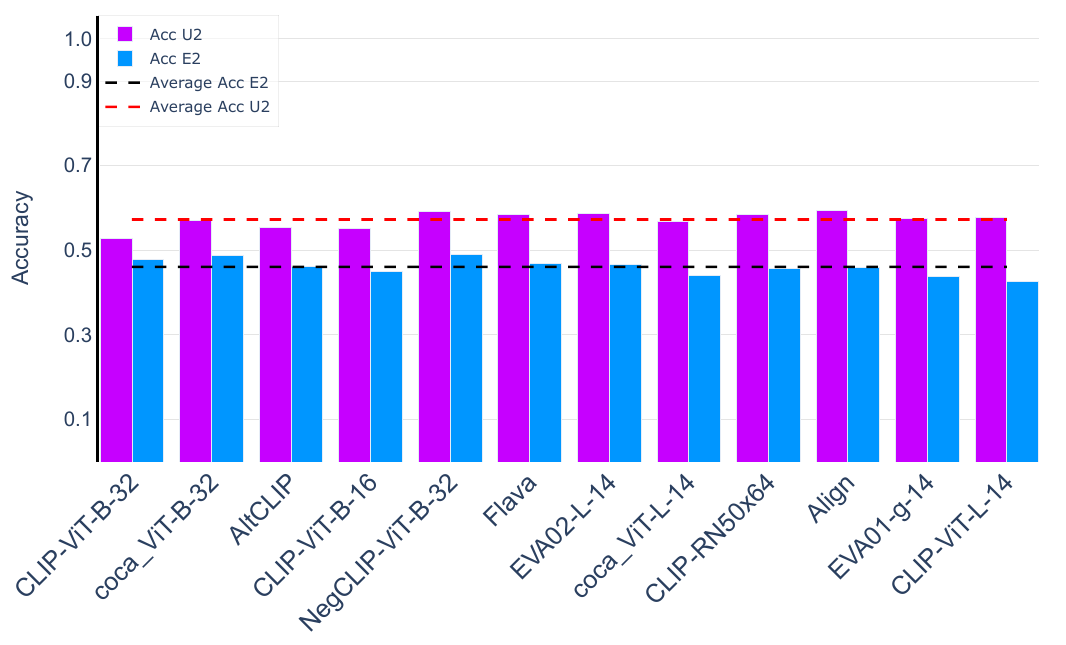}
\caption{Average accuracy of models in the text-to-image retrieval tasks. The images are sourced from the E2 and U2 groups, which are images that feature two individuals of different genders. The performance of models on the U2 and E2 groups is represented by the \textcolor{violet}{purple} and \textcolor{blue}{blue} bars, respectively. Additionally, the \textcolor{red}{red} and \textcolor{black}{black} dashed lines depict the average performance for the U2 and E2 groups, respectively.}
\vspace{1cm}
\label{fig:TextToImage}
\end{figure}

Figure~\ref{fig:TextToImage} illustrates the accuracy of VLMs in the task of text-to-image retrieval. For each activity, we form random pairs of images from E2 and U2\footnote{Pairing is in the form of one-to-one correspondence.}. Each model is then evaluated for its ability to retrieve the image that best matches captions formatted as "a \textless{}man/woman\textgreater{} is \textless{}doing activity\textgreater{} and a \textless{}woman/man\textgreater{} is in the scene". As can be seen in Figure~\ref{fig:TextToImage}, VLMs achieve an accuracy of approximately 50\%, indicating that their performance is nearly random in this task.
If we replicate this experiment, altering only the captions to a gender-neutral phrase "a person is \textless{}doing activity\textgreater{}", and designate the image where the expected gender is performing the activity as the true label, the VLMs achieve an accuracy of approximately 50\% again, as illustrated in Figure 4 in the Appendix.
In essence, the benchmarked models do not seem to comprehend the properties of the given images that would aid in retrieving the correct one given embedded information from the caption, i.e., they fail to recognize the performer of the activity based on the text.

\section{Discussion}\label{sec:discussion}
\subsection{Image and Text Embedding in VLMs}
To gain a deeper understanding of how texts and images are encoded into the shared embedding space, consider these points:
\begin{enumerate}
    \item Given that the image retrieval accuracy is nearly 50\% for both expected and unexpected caption groups (as shown in Figure~\ref{fig:TextToImage}), we can infer that the cosine similarity (as per Eq.~\ref{eq:matching_score}) between both sets of captions and images (for expected and unexpected groups) is almost identical.
    \item As the average text retrieval accuracy exceeds 60\% for E2 and falls below 50\% for U2, it suggests that text embeddings are somewhat closer to the image embeddings in E2.
\end{enumerate}
A plausible interpretation of these observations is that VLMs encode images more closely to the encoding of the more expected text. However, the image encoders of VLMs embed images from both groups closely together, resulting in nearly equal cosine similarity with any caption embedding.

\subsection{Bias Mitigation}
Multiple works have previously discussed approaches for mitigating bias from VLMs, such as orthogonal projection~\citep{chuang2023debiasing}, making unbiased datasets~\citep{li2023debias, howard2024socialcounterfactuals}, representation correction~\citep{seth2023dear,Wang2021AreGQ}, and prompt tuning~\citep{berg2022prompt}. Several of these approaches could be applied also for robustness to gender-activity binding bias. For more detail on these works, refer to the Appendix.

\section{Future Works}
\textbf{Study Other Social Biases.}
In this work, we focused solely on gender bias; however, the experimental approaches described can be extended to other social biases, such as race and age. In future research, we plan to explore how VLMs perform across these dimensions using the experiments outlined in this study.\\
\textbf{Study the Source of Bias in the Training Data.}
Due to limited accessibility to the training sets, especially for models with unpublished datasets, this study could not investigate the source of bias in the training data directly. We plan to address this limitation in future work by examining the sources of bias in publicly available training datasets for VLMs and analyzing the origins of these biases within those datasets.

\section{Conclusions}

In this study, we have explored a significant yet overlooked bias in VLMs known as the gender-activity binding bias. We used a unique dataset named GAB, which includes about 5500 images. We found that the capacity of VLMs to link an activity with the gender of its performer notably decreases when another person of a different gender is in the scene. We discovered that while the bias is evident in image-to-text retrieval tasks and in the text encoder alone, the models perform randomly in text-to-image retrieval tasks. This suggests that the gender-activity binding bias is primarily absorbed by the text encoder rather than the image encoder in VLMs. Furthermore, we noted that while VLMs have difficulty with gender-activity binding, they do have some ability to recognize activities. We believe that the ability of VLMs to comprehend activities and perform compositional reasoning in complex scenes are other crucial factors contributing to the gender-activity binding bias that could be investigated in future research.






\bibliography{mybibfile}

\section{Dataset}
\label{supp:dataset_detail}

\subsection{Detailed Report}
The method of selecting activities is detailed above, ultimately resulting in three categories of biased activities. comprehensive report of each category provided in Table~\ref{table:complete_rel_ster1} , Table~\ref{table:complete_rel_ster2} , Table~\ref{table:complete_rel_ster3}.



\begin{table}[ht!]
\centering
\caption{List of all gender-biased \textit{stereotypical} activities along with their expected subject and amount of data in each setting for them.}
\vspace{1cm}
\resizebox{0.5\textwidth}{!}{%
\begin{tabular}{lccccc}\toprule
\multirow{2}{*}{Activity} & \multirow{2}{*}{Exp. Subj.} & \multicolumn{4}{c}{Setting} \\\cmidrule(lr){3-6}
& & E1 & E2 & U1 & U2\\
\midrule
Applying Nail Polish & Women & 43 & 35 & 43 & 45\\
Carving The Turkey & Men & 38 & 48 & 41 & 30 \\
Crafting Homemade Gifts & Women & 41 & 28 & 41 & 43\\
Doing Pilates & Women & 40 & 34 & 37 & 32 \\
Handling Barbecue Equipment & Men & 47 & 55 & 48 & 35\\
Participating in American Football & Men & 44 & 34 & 39 & 30\\
Performing Physical Repairs & Men& 38 & 42 & 41 & 34 \\
Playing Poker & Men & 41  & 54 & 38 & 34 \\
Preparing Herbal Teas & Women& 43 & 33 & 55 & 38\\
Shaving Facial Hair & Men & 44 & 38 & 39 & 32\\
Tying Child's Hair into a Ponytail & Women & 38 & 30 & 29 & 44\\
Unclogging Drains or Fixing Leaks & Men & 36 & 44 & 35 & 27\\
\bottomrule
\end{tabular}}
\label{table:complete_rel_ster1}
\end{table}



\begin{table}[ht!]
\centering
\caption{List of all gender-biased everyday activities along with their expected subject and amount of data in each setting for them.}
\vspace{1cm}
\resizebox{0.5\textwidth}{!}{%
\begin{tabular}{lccccc}\toprule
\multirow{2}{*}{Activity} & \multirow{2}{*}{Exp. Subj.} & \multicolumn{4}{c}{Setting} \\\cmidrule(lr){3-6}
& & E1 & E2 & U1 & U2\\
\midrule
Baking Cake & Women & 40 & 33 & 42 & 35\\
Customizing Motorcycle & Men & 48 & 46 & 49 & 31\\
Designing Building & Men & 37 & 40 & 37 & 36\\
Designing Dress & Women & 56 & 31 & 49 & 41\\
Diving From High Board & Men & 39 & 38 & 30 & 28\\
Doing Basket Weaving & Women & 41 & 30 & 45 & 40\\
Doing Professional Makeup & Women & 38 & 40 & 28 & 36\\
Having Face Mask & Women & 37 & 36 & 43 & 40\\
Knitting Colorful Scarves & Women & 42 & 37 & 42 & 32\\
Playing Acoustic Guitar & Men & 43 & 46 & 37 & 37\\
Programming New Software & Men & 59 & 64 & 46 & 27\\
Repairing Electronics & Men & 42 & 41 & 39 & 41\\
Skating In a Roller Derby & Women & 44 & 30 & 33 & 42\\
Spinning Pottery & Women & 61 & 31 & 38 & 59\\
Tuning Guitar & Men & 39 & 41 & 39 & 42\\
\bottomrule
\end{tabular}}
\label{table:complete_rel_ster2}
\end{table}



\begin{table}[ht!]
\centering
\caption{List of all gender-biased activities from LAION-400M along with their expected subject and amount of data in each setting for them.}
\vspace{1cm}
\resizebox{0.5\textwidth}{!}{%
\begin{tabular}{lccccc}
\toprule
\multirow{2}{*}{Activity} & \multirow{2}{*}{Exp. Subj.} & \multicolumn{4}{c}{Setting} \\\cmidrule(lr){3-6}
& & E1 & E2 & U1 & U2\\
\midrule
Baking Bread & Women & 44 & 26 & 33 & 31\\
Beading Earrings & Women & 33 & 38 & 44 & 46\\
Catching Fish & Men & 36 & 32 & 54 & 31\\
Choosing Dress & Women & 28 & 29 & 30 & 42\\
Climbing Tree & Men & 27 & 28 & 34 & 42\\
Drinking Beer & Men & 31 & 31 & 101 & 41\\
Holding Baby & Women & 54 & 31 & 31 & 44\\
Holding Gun & Men & 32 & 32 & 42 & 37\\
Leading Team & Men & 31 & 29 & 32 & 30\\
Picking Flower & Women & 31 & 31 & 34 & 32\\
\bottomrule
\end{tabular}}
\label{table:complete_rel_ster3}
\end{table}

\subsection{Experimented Models For Image Generation}
\label{app:generationAppendix}
Initially, we explored several diffusion\cite{ho2020denoising} models including the original Stable Diffusion\cite{rombach2022highresolution}, Stable Diffusion XL\cite{podell2023sdxl}, and its checkpoints like (Juggernaut XL\footnote{https://civitai.com/models/133005/juggernaut-xl} and DreamShaper XL\footnote{https://civitai.com/models/112902/dreamshaper-xl}). We tried Fooocus\footnote{https://github.com/lllyasviel/Fooocus}, an image generation tool that includes its prompt rewriting system using GPT-2\cite{radford2019language}. The generated images were realistic and diverse; however, they lacked representation of the intended activity, especially in the context of compositional generation. Upon using DALL-E 3 with OpenAI API, we observed a significant improvement in the quality of compositional generation while also keeping the reality and the clarity of the image. We chose DALL-E 3 as our final image generation solution.

\subsection{Used LLM For Prompt Enhancement}
\label{app:enhancementAppendix}
We evaluated various large language models (LLMs) for this task. Due to our hardware constraints, we ultimately selected the Llama2 model, which has 70 billion parameters, to generate prompts \citep{touvron2023llama}.

\section{Bias Mitigation}
Multiple works have previously discussed approaches for mitigating bias from VLMs that could be applied also for robustness to gender-activity binding bias. 

In \cite{chuang2023debiasing}, an orthogonal projection is applied to the end embeddings, making them orthogonal to the embeddings of spurious prompts, effectively mitigating bias introduced by these prompts. This alteration might also strip away useful nuances from the embeddings, potentially degrading the model’s overall performance on tasks where those nuances are beneficial and also this might not hold true in complex real-world scenarios where biases are not strictly linear.

In \cite{seth2023dear}, a method is introduced that utilizes additive residual image representations to correct biases in the original representations. However, training these residuals demands extensive and diverse data, making the process resource-intensive. Moreover, incorporating these residuals may alter other inherent characteristics of the original representations, potentially affecting the overall performance of the model.

In \cite{berg2022prompt}, learnable prompts integrated into the text input are used to mitigate bias, optimized through the training of an adversarial classifier that evaluates similarity scores between outputs from two encoders. This method employs both adversarial and contrastive losses to preserve the quality of the joint representation while aiming to reduce bias. However, the effectiveness of this adversarial training depends on the availability of a large and diverse dataset; otherwise, there is a risk of overfitting to adversarial objectives. Additionally, since VLMs are trained with various objective functions, making adjustments to mitigate bias could potentially lead to trade-offs in other performance metrics.

Both \citet{howard2024socialcounterfactuals} and \citet{li2023debias} address the creation of fair datasets to mitigate bias in VLMs, though each method has its limitations. \citet{howard2024socialcounterfactuals} uses AI to generate datasets for fine-tuning VLMs, which is computationally intensive and can introduce new biases due to the limitations of AI generation tools. On the other hand, \citet{li2023debias} employs counterfactual data—images and texts modified to alter specific attributes. While this method targets specific biases directly, it is limited to those attributes it explicitly modifies and may not adequately address all potential biases.

In \citet{Wang2021AreGQ}, the authors address bias by calculating the mutual information between each dimension of the encoder outputs and the gender attribute, clipping dimensions with the highest mutual information to reduce their influence. While this approach effectively reduces gender bias in embeddings, it faces limitations similar to other inference-time interventions. Notably, it may not adapt well to a wide range of biases and could potentially degrade the accuracy of models in other downstream tasks due to the removal of critical information during the clipping process.

\section{Results}

\subsection{Image-to-Text Retrieval}\label{app:ittr}

\begin{figure}[ht!]
\centering
\includegraphics[width=0.48\textwidth]{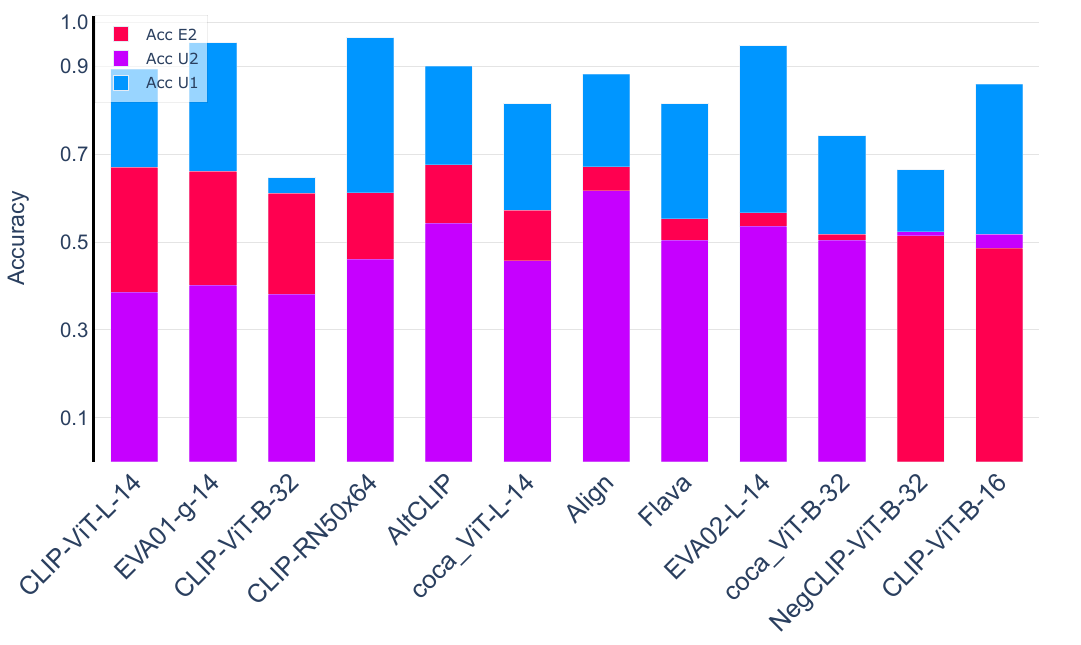}
\caption{Average retrieval accuracy of VLMs on the image-to-text retrieval task across various scenarios. The chart highlights the performance drop between these scenarios for each model.(In this chart, the performance of the models on gender-biased activities from LAION-400M is reported, with the experiment is similar to the one described in Figure 2 of the main text.)}
\vspace{1cm}
\label{fig:phaze1}
\end{figure}

\begin{figure}[ht!]
\centering
\includegraphics[width=0.48\textwidth]{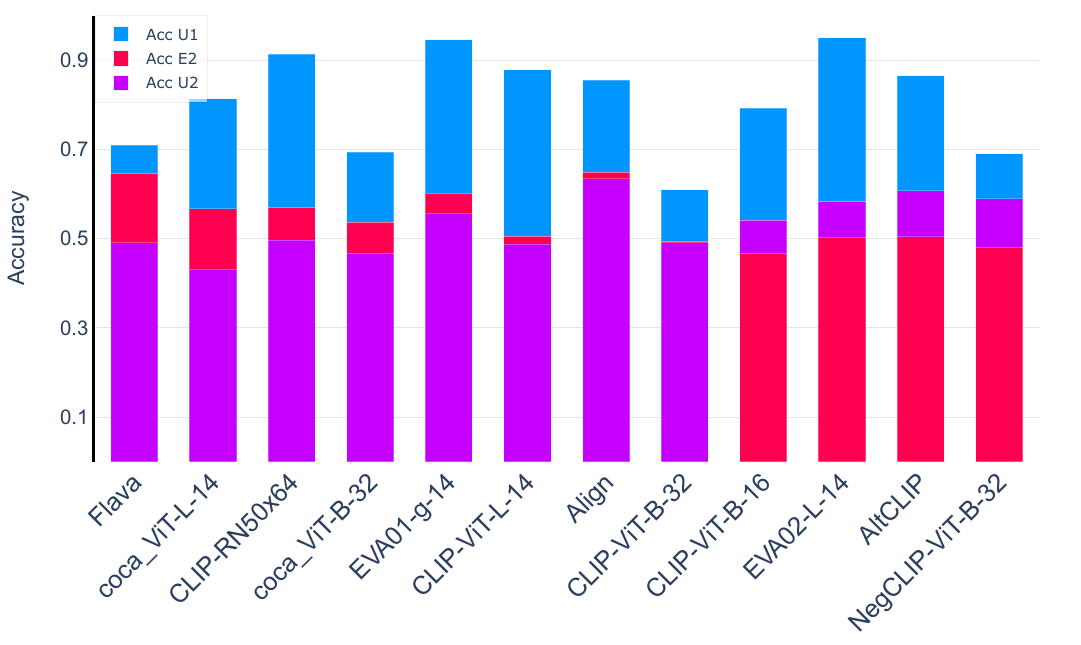}
\caption{Average retrieval accuracy of VLMs on the image-to-text retrieval task across various scenarios. The chart highlights the performance drop between these scenarios for each model.(In this chart, the performance of the models on gender-biased everyday activities is reported, with the experiment is similar to the one described in Figure 2 of the main text.)}
\vspace{1cm}
\label{fig:phaze2}
\end{figure}

The performance of text retrieval for texts featuring two genders (one of them performing the activity) is demonstrated in Table~\ref{fig:phaze1} and Table~\ref{fig:phaze2}. This is akin to the experiment discussed in Section 5.2.1,
In this experiment, we assessed image-to-text retrieval across three distinct scenarios. The first scenario involves images where an individual of an unexpected gender is depicted performing a typically biased activity with no other individuals present in the scene. The second scenario comprises images where both genders are present, and the activity is performed by the gender typically associated with it. The third scenario mirrors the second, but in this case, the activity is performed by the gender not typically associated with it. In Table~\ref{fig:phaze1} the experiment is reported for gender-biased activities from LAION-400M and in Table~\ref{fig:phaze2} the experiment is reported for gender-biased everyday activities.

\subsection{Text-to-Image Retrieval}\label{app:ttir}

\begin{figure}[ht!]
\centering
\includegraphics[width=0.48\textwidth]{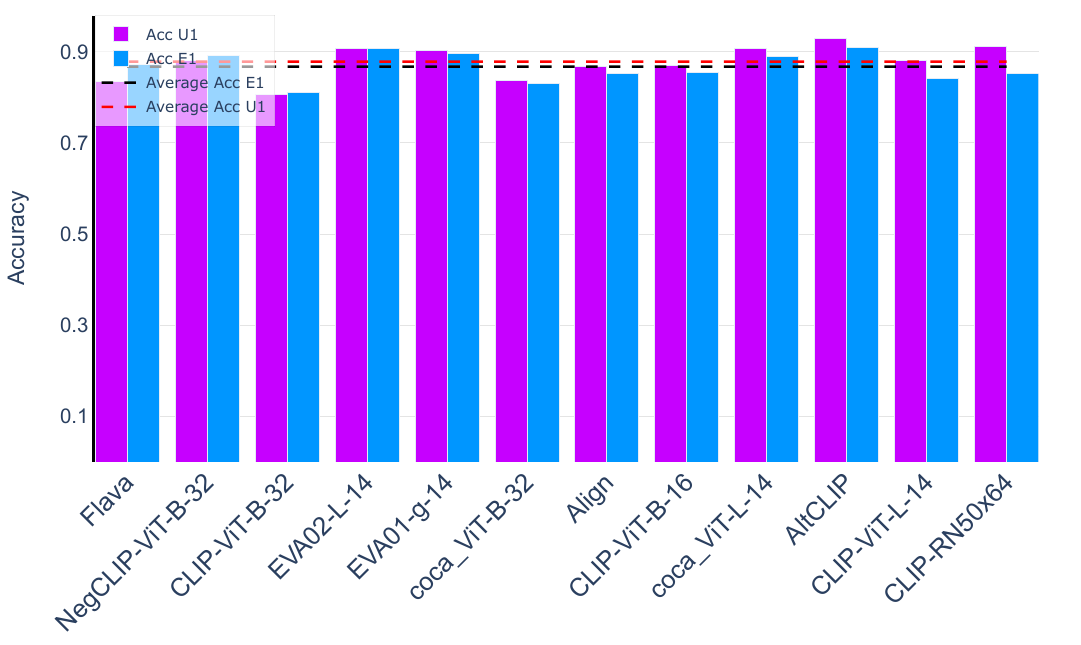}
\caption{Average accuracy of models in the text-to-image retrieval tasks. The images are sourced from the E1 and U1 groups, which are images that feature one individual gender. The performance of models on the U1 and E1 groups is represented by the \textcolor{violet}{purple} and \textcolor{blue}{blue} bars, respectively. Additionally, the \textcolor{red}{red} and \textcolor{black}{black} dashed lines depict the average performance for the U1 and E1 groups, respectively (The experiment is similar to the one described in Figure 3 of the main text.).}
\vspace{1cm}
\label{fig:text_to_image_single}
\end{figure}

\begin{figure}[ht!]
\centering
\includegraphics[width=0.4\textwidth]{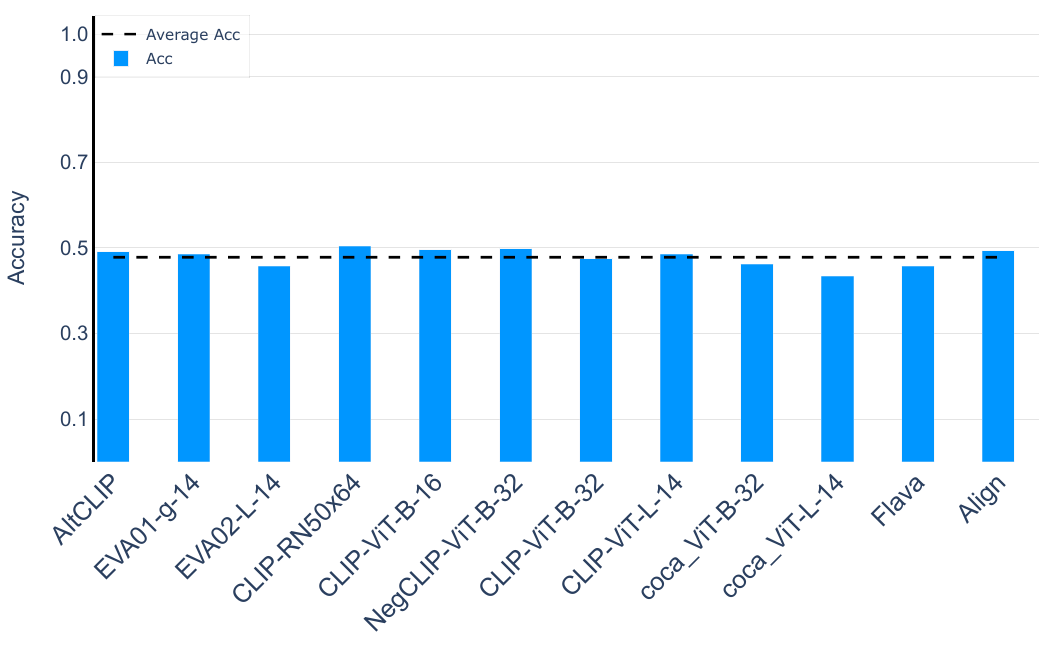}
\caption{The percentage of all activities in each model, where the embedding of the sentence “a person is \textless{}doing activity\textgreater{}” has a higher matching score (Eq.3 of the main text) with the images of the E1 group than those of the U1 group (defined in Section 4 in the main text. ).}
\vspace{1cm}
\label{fig:Image_bias_single}
\end{figure}

The performance of image retrieval for images featuring a single individual (the one performing the activity) is demonstrated in Figure \ref{fig:text_to_image_single}. This is akin to the experiment discussed in Section 5.4 of the main text, where images in E1/U1 are categorized and the models are evaluated based on their ability to accurately retrieve the correct image using captions that follow the template “a \textless{}man/woman\textgreater{} is \textless{}performing an activity\textgreater{}”. The models exhibit an average accuracy of approximately 90\%, indicating that they can effectively identify the image that correctly represents the gender of the individual performing the activity, provided there are no individuals of a different gender present in the scene.
Also, you can see the results of experiment mentioned in Section 5.4 of the main text in Figure \ref{fig:Image_bias_single}.

\subsection{Activity Recognition}
\label{activity recognition}
A crucial inquiry concerning the bias in binding gender with certain activities is whether this bias originates from a misunderstanding of the activity itself, and if this misunderstanding leads to a failure in binding the gender and the activity. To gauge the model’s understanding of activities, we collect several (at least 9, 17.25 on average) other different activities that are performed in the same environment as the reference activity for each activity in the GAB dataset using GPT-4~\citep{openai2023gpt4}. For each image in the GAB dataset, we create a caption from the reference and associate collected activities and arrange them based on their matching score to the image. We show the mean reciprocal rank (MRR), recall@1, and recall@3 for the true caption (reference activity) in Table~\ref{table:metrics_categorized} for stereotypical activities. (results for other categories of activities is depicted in Table~\ref{table:metrics_categorized_phaze2} and Table~\ref{table:metrics_categorized_phaze1}).

The table indicates that the model can, to some extent, retrieve the true activity from a list of activities performed in a similar environment, demonstrating that VLMs can recognize the performed activity in image-to-text retrieval.


\begin{table}[ht]
\centering
\caption{Models' ability to retrieve the correct activity from a list of activities. For every image, we formulate a collection of captions derived from a list of activities and arrange them according to their matching score (Eq.3 of the main text) with the image. We report metrics such as Mean Reciprocal Rank (MRR), recall@1, and recall@3, which indicate that VLMs can to some degree recognize the correct captions in image-to-text retrieval tasks.}
\vspace{1cm}
\resizebox{0.5\textwidth}{!}{%
\begin{tabular}{lcccccc}
\toprule
\textbf{Model} & \multicolumn{2}{c}{\textbf{MRR}} & \multicolumn{2}{c}{\textbf{Recall@1}} & \multicolumn{2}{c}{\textbf{Recall@3}} \\
\cmidrule(lr){2-3} \cmidrule(lr){4-5} \cmidrule(lr){6-7}
               & \textbf{One Gender} & \textbf{Two Genders} & \textbf{One Gender} & \textbf{Two Genders} & \textbf{One Gender} & \textbf{Two Genders} \\ & \textbf{Presence} & \textbf{Presence} & \textbf{Presence} & \textbf{Presence} & \textbf{Presence} & \textbf{Presence}\\
\midrule
AltCLIP & 0.464 & 0.431 & 0.311 & 0.268 & 0.511 & 0.477 \\
EVA01-g-14  & 0.448 & 0.347 & 0.267 & 0.165 & 0.517 & 0.395 \\
EVA02-L-14  &0.565 & 0.493 & 0.391 & 0.307 & 0.668 & 0.583 \\
CLIP-RN50x64 & 0.423 & 0.437 & 0.259 & 0.277 & 0.467 & 0.482 \\
CLIP-ViT-B-16 & 0.346 & 0.334 & 0.187 & 0.199 & 0.378 & 0.347 \\
NegCLIP-ViT-B-32 & 0.292 & 0.287 & 0.124 & 0.128 & 0.305 & 0.305 \\
CLIP-ViT-B-32 & 0.256 & 0.267 & 0.087 & 0.102 & 0.255 & 0.278 \\
CLIP-ViT-L-14 & 0.387 & 0.368 & 0.215 & 0.169 & 0.414 & 0.450 \\
COCA-ViT-B-32 & 0.331 & 0.341 & 0.179 & 0.192 & 0.350 & 0.351 \\
COCA-ViT-L-14 & 0.403 & 0.378 & 0.240 & 0.209 & 0.434 & 0.425 \\
Flava & 0.404 & 0.388 & 0.216 & 0.204 & 0.465 & 0.442 \\
Align & 0.384 & 0.360 & 0.220 & 0.176 & 0.419 & 0.407 \\
\bottomrule
\end{tabular}}
\label{table:metrics_categorized}
\end{table}

\begin{table}[ht]
\centering
\caption{In this table, the reported metrics indicate the models’
ability to retrieve the correct activity from among those gathered
from GPT-4. (In this chart, the performance of the models on gender-biased everyday activities is reported, with the experiment is similar to the one described in Table \ref{table:metrics_categorized}.)}
\vspace{1cm}
\label{table:metrics_categorized_phaze2}
\resizebox{0.5\textwidth}{!}{%
\begin{tabular}{lcccccc}
\toprule
\textbf{Model} & \multicolumn{2}{c}{\textbf{MRR}} & \multicolumn{2}{c}{\textbf{Recall@1}} & \multicolumn{2}{c}{\textbf{Recall@3}} \\
\cmidrule(lr){2-3} \cmidrule(lr){4-5} \cmidrule(lr){6-7}
               & \textbf{One Gender} & \textbf{Two Genders} & \textbf{One Gender} & \textbf{Two Genders} & \textbf{One Gender} & \textbf{Two Genders} \\ & \textbf{Presence} & \textbf{Presence} & \textbf{Presence} & \textbf{Presence} & \textbf{Presence} & \textbf{Presence}\\
\midrule
AltCLIP         & 0.677 & 0.593 & 0.553 & 0.449 & 0.731  & 0.659 \\
EVA01-g-14      & 0.585 & 0.512 & 0.425 & 0.363 & 0.672 & 0.564 \\
EVA02-L-14      & 0.688 & 0.563 & 0.564 & 0.415 & 0.760 & 0.635 \\
CLIP-RN50x64    & 0.601 & 0.495 & 0.465 & 0.334 & 0.648 & 0.557 \\
CLIP-ViT-B-16   & 0.483 & 0.456 & 0.333 & 0.310 & 0.531 & 0.486 \\
NegCLIP-ViT-B-32& 0.347 & 0.332 & 0.159 & 0.153 & 0.414 & 0.385  \\
CLIP-ViT-B-32   & 0.296 & 0.269 & 0.128 & 0.110 & 0.316  & 0.274 \\
CLIP-ViT-L-14   & 0.606 & 0.550 & 0.469 & 0.404 & 0.672 & 0.618 \\
COCA-ViT-B-32   & 0.319 & 0.292 & 0.137 & 0.128 & 0.382 & 0.316 \\
COCA-ViT-L-14   & 0.459 & 0.418 & 0.280  & 0.243 & 0.529 & 0.489  \\
Flava           & 0.474 & 0.469 & 0.283  & 0.282 & 0.554 & 0.538 \\
Align           & 0.534 & 0.497 & 0.375 & 0.331 & 0.620 & 0.589 \\
\bottomrule
\end{tabular}}
\end{table}

\begin{table}[ht]
\centering
\caption{In this table, the reported metrics indicate the models’ ability to retrieve the correct activity from among those gathered from GPT-4. (In this chart, the performance of the models on gender-biased activities from LAION-400M is reported, with the experiment is similar to the one described in Table \ref{table:metrics_categorized}.)}
\vspace{1cm}
\label{table:metrics_categorized_phaze1}
\resizebox{0.5\textwidth}{!}{%
\begin{tabular}{lcccccc}
\toprule
\textbf{Model} & \multicolumn{2}{c}{\textbf{MRR}} & \multicolumn{2}{c}{\textbf{Recall@1}} & \multicolumn{2}{c}{\textbf{Recall@3}} \\
\cmidrule(lr){2-3} \cmidrule(lr){4-5} \cmidrule(lr){6-7}
               & \textbf{One Gender} & \textbf{Two Genders} & \textbf{One Gender} & \textbf{Two Genders} & \textbf{One Gender} & \textbf{Two Genders} \\ & \textbf{Presence} & \textbf{Presence} & \textbf{Presence} & \textbf{Presence} & \textbf{Presence} & \textbf{Presence}\\
\midrule
AltCLIP        & 0.561 & 0.456 & 0.426 & 0.295 & 0.628 & 0.527 \\
EVA01-g-14     & 0.550 & 0.409 & 0.414 & 0.235 & 0.607 & 0.478 \\
EVA02-L-14     & 0.696 & 0.563 & 0.592 & 0.418  & 0.757 & 0.627 \\
CLIP-RN50x64   & 0.568 & 0.467 & 0.437 & 0.318 & 0.620 & 0.505 \\
CLIP-ViT-B-16  & 0.440 & 0.389 & 0.325 & 0.265 & 0.438 & 0.391  \\
NegCLIP-ViT-B-32 & 0.327 & 0.273 & 0.169 & 0.132 & 0.374 & 0.282 \\
CLIP-ViT-B-32  & 0.413  & 0.379 & 0.268 & 0.240 & 0.455 & 0.410  \\
CLIP-ViT-L-14  & 0.454 & 0.372 & 0.300 & 0.207  & 0.518 & 0.432 \\
COCA-ViT-B-32  & 0.239 & 0.232 & 0.080  & 0.078 & 0.237 & 0.204 \\
COCA-ViT-L-14  & 0.449 & 0.369 & 0.316  & 0.202 & 0.477 & 0.412 \\
Flava          & 0.384 & 0.375 & 0.207 & 0.173 & 0.433  & 0.440 \\
Align          & 0.571 & 0.502 & 0.429 & 0.350 & 0.636 & 0.551 \\
\bottomrule
\end{tabular}}
\end{table}

\end{document}